\documentclass[11pt,a4paper]{article}
\usepackage{authblk}
\usepackage[hyperref]{acl2020}
\usepackage{graphicx}
\usepackage{times}
\usepackage{latexsym}
\usepackage{float}
\usepackage{subfigure}
\usepackage{amsmath}
\usepackage{amssymb}
\usepackage{multirow}
\usepackage{booktabs}
\usepackage{algorithm}
\usepackage{algorithmic}
\usepackage{bm}

\newcommand{\tabincell}[2]{\begin{tabular}{@{}#1@{}}#2\end{tabular}}

\usepackage{microtype}

\aclfinalcopy 

\title{Ultra-Fast, Low-Storage, Highly Effective Coarse-grained Selection in Retrieval-based Chatbot by Using Deep Semantic Hashing}

\makeatletter
\newcommand\email[2][]%
   {\newaffiltrue\let\AB@blk@and\AB@pand
      \if\relax#1\relax\def\AB@note{\AB@thenote}\else\def\AB@note{\relax}%
        \setcounter{Maxaffil}{0}\fi
      \begingroup
        \let\protect\@unexpandable@protect
        \def\thanks{\protect\thanks}\def\footnote{\protect\footnote}%
        \@temptokena=\expandafter{\AB@authors}%
        {\def\\{\protect\\\protect\Affilfont}\xdef\AB@temp{#2}}%
         \xdef\AB@authors{\the\@temptokena\AB@las\AB@au@str
         \protect\\[\affilsep]\protect\Affilfont\AB@temp}%
         \gdef\AB@las{}\gdef\AB@au@str{}%
        {\def\\{, \ignorespaces}\xdef\AB@temp{#2}}%
        \@temptokena=\expandafter{\AB@affillist}%
        \xdef\AB@affillist{\the\@temptokena \AB@affilsep
          \AB@affilnote{}\protect\Affilfont\AB@temp}%
      \endgroup
       \let\AB@affilsep\AB@affilsepx
}
\makeatother

\author[1]{\textbf{Tian Lan}}
\author[1]{\textbf{Xian-Ling Mao}}
\author[1]{\textbf{Xiaoyan Gao}}
\author[2]{\textbf{Wei Wei}}
\author[1]{\textbf{Heyan Huang}}
\affil[1]{Beijing Institute of Technology}
\email{\url{lantiangmftby@gmail.com},\url{{maoxl,xygao,hhy63}@bit.edu.cn}}
\affil[2]{Huazhong University of Science and Technology}
\email{\url{Weiw@hust.edu.cn}}

\date{}

\begin{document}
\maketitle

\begin{abstract}
  We study the coarse-grained selection module in retrieval-based chatbot.
  Coarse-grained selection is a basic module in a retrieval-based chatbot,
  which constructs a rough candidate set from the whole database to speed up the interaction with customers.
  So far, there are two kinds of approaches for coarse-grained selection module: 
  (1) sparse representation; (2) dense representation.
  To the best of our knowledge, there is no systematic comparison between these two approaches in retrieval-based chatbots,
  and which kind of method is better in real scenarios is still an open question.
  In this paper, we first systematically compare these two methods from four aspects: 
  (1) effectiveness; (2) index stoarge; (3) search time cost; (4) human evaluation.
  Extensive experiment results demonstrate that dense representation method 
  significantly outperforms the sparse representation, 
  but costs more time and storage occupation.
  In order to overcome these fatal weaknesses of dense representation method, 
  we propose an ultra-fast, low-storage, and highly effective 
  \textbf{D}eep \textbf{S}emantic \textbf{H}ashing \textbf{C}oarse-grained selection method, called DSHC model.
  Specifically, in our proposed DSHC model,
  a hashing optimizing module that consists of two autoencoder models is 
  stacked on a trained dense representation model,
  and three loss functions are designed to optimize it.
  The hash codes provided by hashing optimizing module effectively 
  preserve the rich semantic and similarity information in dense vectors.
  Extensive experiment results prove that,
  our proposed DSHC model can achieve much faster speed and lower storage than sparse representation,
  with limited performance loss compared with dense representation.
  Besides, our source codes have been publicly released for future research\footnote{\url{https://github.com/gmftbyGMFTBY/HashRetrieval}}.
\end{abstract}

\section{Introduction}

Retrieval technique or response selection is a very popular and elegant approach 
to framing a chatbot i.e. open-domain dialog system.
Given the conversation context, a retrieval-based chatbot aims to select the most appropriate utterance as a response
from a pre-constructed database. 
In order to balance the effectiveness and efficiency,
mosts of the retrieval-based chatbots \cite{Fu2020ContexttoSessionMU} employ
coarse-grained selection module to recall a set of candidate 
that are semantic coherent with the conversation context to speed up processing.

To the best of our knowledge, there are two kinds of approaches 
to build a coarse-grained selection module in retrieval-based chatbots:
(1) sparse representation: 
TF-IDF or BM25 \cite{Robertson2009ThePR} is a widely used method.
It matches keywords with an inverted index and can be seen
as representing utterances in highdimensional sparse vectors \cite{Karpukhin2020DensePR};
(2) dense representation: 
Large scale pre-trained langauge models (PLMs), e.g. BERT \cite{Devlin2019BERTPO} are
commonly used to obtain the semantic representation of utterances,
which could be used to recall semantic coherent candidates by using cosine similarity \cite{Karpukhin2020DensePR}.

So far, there is no systematic comparison between these two kinds of approaches in retrieval-based chatbots,
and which kind of method is most appropriate in real scenarios is 
still an open question that confuses researchers in dialog system community.
Thus, in this paper, we first conduct extensive experiment to compare these two approaches
from four important aspects: 
(1) effectiveness; (2) search time cost; (3) index storage occupation; (4) human evaluation.
Extensive experiment results on four popular response selection datasets 
demonstrate that the dense representation 
significantly outperforms the sparse representation at the expense of 
the lower speed and bigger storage than sparse representation,
which is unsufferable in real scenarios.
Then, in order to overcome the fatal weaknesses of dense representation methods,
we propose an ultra-fast, low-storage and highly effective 
\textbf{D}eep \textbf{S}emantic \textbf{H}ashing \textbf{C}oarse-grained selection module (DSHC)
which effectively balances the effectiveness and efficiency.
Specifically, 
we first stack a novel hashing optimizing module that consists of two autoencoders on a given 
dense representation method.
Then, three well designed loss functions are used to optimize 
these two autoencoders in hashing optimizing module:
(1) preserved loss; (2) hash loss; (3) quantization loss.
After training, the autoencoders could effectively preserve rich semantic and similarity information
of the dense vectors into the hash codes,
which are very computational and storage efficient \cite{Wang2018ASO}.
Extensive experiment results on four popular response selection datasets
demonstrate that our proposed DSHC model can achieve much faster search speed 
and lower storage occupation than sparse representation method,
and very limited performance loss compared with the given dense representation method.

In this paper, our contributions are three-fold:
\begin{itemize}
  \item We systematically compare current two kinds of coarse-grained selection methods in open-domain retrieval-based dialog systems 
  from four important aspects: (1) effectiveness; (2) search time cose; (3) stoarge occupation; (4) human evaluation.
  \item We propose an ultra-fast, low-storage, and highly effective 
  deep semantic hashing coarse-grained selection method, called DSHC, 
  which overcomes the fatal weaknesses of the dense representation method.
  \item We have publicly released our source codes for future search.
\end{itemize}

The rest of this paper is organized as follows:
we introduce the important concepts and background covered in our paper in Section 2.
The experiment settings is presented in Section 3.
In Section 4, we systematically compare the current two kinds of methods in coarse-grained selection module: 
(1) sparse representation; (2) dense representation.
In Section 5, we introduce our proposed DSHC model, and detailed experiment results are elaborated.
In Section 6, we conduct the case study.
Finally, we conclude our work in Section 7.
Due to the page limitation, more details and extra analysis can be found in \textit{Appendix}.

\section{Preliminary}
\subsection{Retrieval-based Chatbot}
Retrieval-based chatbots, or retrieval-based open-domain dialog systems,
which are widely used in the real scenarios,
have gained great progress over the past few years.
So far, most of the retrieval-based chatbots contain two modules \cite{Fu2020ContexttoSessionMU,Luan2020SparseDA}:
coarse-grained selection module and fine-graied selection module.

\subsubsection{Coarse-grained Selection Module}

Coarse-grained selection module recalls a set of candidate responses that are semantic coherent with the conversation context 
from the pre-constructed database.
As described before, there are two kinds of approaches to construct a coarse-grained selection module:
sparse and dense representation.

\textbf{Sparse representation:}
Due to the simply implementation and effective performance,
sparse representation methods, represented by TF-IDF and BM25 \cite{Robertson2009ThePR},
have been widely used in lots of real applications.
Because the utterance that has the keywords overlap with the conversation context 
is likely to be an appropriate candidate response,
sparse representation could effectively recall appropriate candidates for the fine-grained selection module.

The advantage of this method is that it runs very quickly.
As shown in Table \ref{tab:6}, with the help of the well designed data structure, 
such as inverted index and skiplist,
it can achieve the best computational complexity $O(\log n)$.
However, there are still lots of appropriate candidate responses 
that don't have the word overlap with the conversation context,
but have very high semantic correlation with the context.
Sparse representation cannot effectively find these cases in the pre-constructed database, 
which may lead to the bad performance.
For example, as shown in Table \ref{tab:1},
it can be found that, the ratio of the ground-truths that can be retrieved by considering word-overlap is low.

\begin{figure}[h]     
  \center{\includegraphics[width=0.5\textwidth, height=4.5cm] {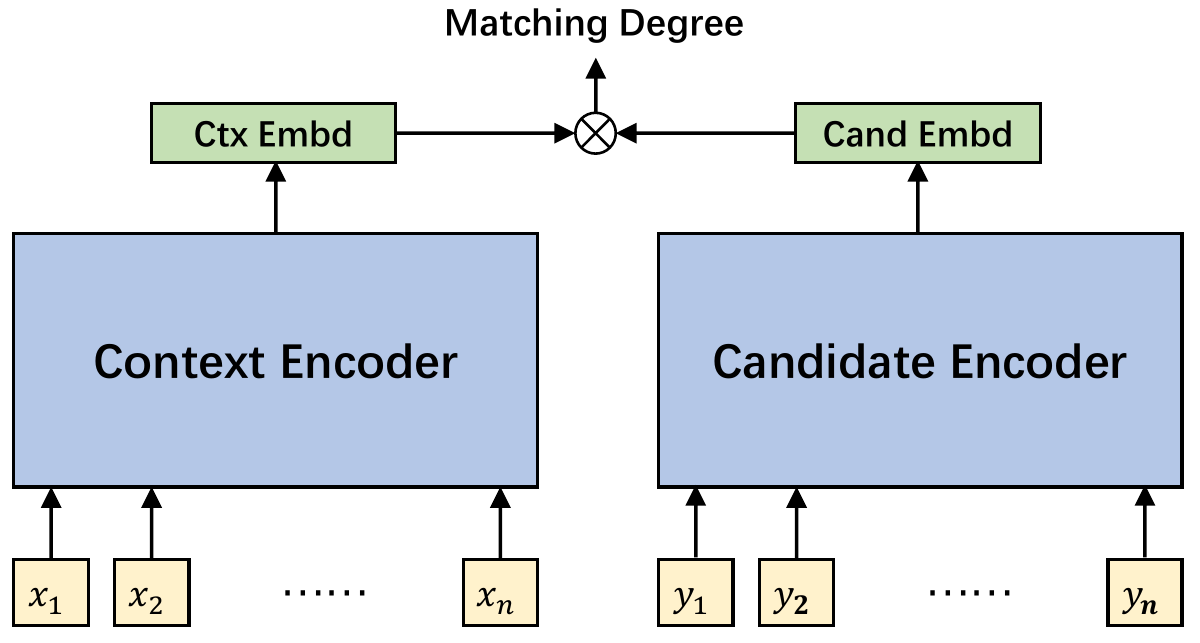}}        
  \caption{The dual-encoder architecture for coarse-grained selection module. $x_i$ and $y_i$ are the $i$th token in conversation context and candidate.
  The matching degree can be obtained by using dot production \cite{Humeau2020PolyencodersAA}.}
  \label{img:1}
\end{figure}

\textbf{Dense representation}:
Recently, dense representation methods, represented by dual-encoder architecture,
\cite{Lowe2015TheUD,Tahami2020DistillingKF,Humeau2020PolyencodersAA,Karpukhin2020DensePR},
have attracted increasing attention of researchers, 
because the rich semantic information could be effectively leveraged.
Besides, large scale pre-trained language models (PLMs) significantly boost the performance of dense representation methods.
As shown in Figure \ref{img:1}, it can be found that,
a dense representation method that leverages the dual-encoder architecture
contains two modules: 
(1) \textbf{semantic encoders} \cite{Humeau2020PolyencodersAA,Tahami2020DistillingKF,Karpukhin2020DensePR} 
are used to obtain the semantic representations of context and candidate responses.
It should be noted that, context semantic encoder and candidate semantic encoder don't share the parameters,
and are optimized separately during training;
(2) \textbf{matching degree} is calculated by using dot production or cosine similarity,
and utterances that have Top-K matching degrees will be selected as the candidates.

However, due to the high computational burden of similarity calculating,
dense representation method runs very slowly.
As shown in Table \ref{tab:6}, 
it can be found that its computational complexity is much bigger than sparse representation methods.

\subsubsection{Fine-grained Selection Module}

Based on the candidate responses provided by the coarse-grained selection module,
fine-grained selection module selects the most appropriate one 
as the final response to the given conversation context.
Over the past few years, there are numerous works proposed to improve the performance of 
fine-grained selection module in retrieval-based chatbots 
\cite{Wu2017SequentialMN,Zhang2018ModelingMC,Zhou2018MultiTurnRS,Tao2019MultiRepresentationFN,Gu2019InteractiveMN,Tao2019OneTO,Yuan2019MultihopSN}.
Especially, recent works \cite{Whang2019DomainAT,Gu2020SpeakerAwareBF} achieve the state-of-the-art results for fine-grained selection by using large scale pre-trained language models (PLMs), e.g. BERT \cite{Devlin2019BERTPO}.
However, because of the diminishing returns \cite{Bisk2020ExperienceGL},
it becomes more and more difficult to improve the open-domain dialog systems by updating the fine-grained selection module.
Compared with the fine-grained selection module, 
there are very few works to study the coarse-grained selection module,
which is a potential breakthrough to improve the retrieval-based open-domain dialog systems further,
and ignored by most of works.

In this paper, a fine-grained selection module serves two purposes:
(1) Construct a reliable metric that measures the average correlation between the conversation context and candidates;
(2) Build retrieval-based chatbots with different coarse-grained selection modules 
to measure their whole performances.

\subsection{Deep Semantic Hashing}

Due to computational and storage efficiencies of the compact binary hash codes,
hashing methods has been widely used for large-scale similarity search \cite{Xu2015ConvolutionalNN}.
The main methodology of deep hashing is similarity preserving, 
i.e., minimizing the gap between the similarities computed in the original space and the similarities
in the hashing code space \cite{Wang2018ASO}.
After optimizing, the hashing codes could save the rich semantic information and similarity information
in original dense vectors.

\section{Experiment Settings}
\subsection{Datasets}
In this paper, we select four popular chinese open-domain dialog datasets:
\begin{itemize}
  \item \textbf{E-Commerce Corpus} \cite{Zhang2018ModelingMC} is collected from the real world conversations between the customers and the service staff 
  from the largest ecommerce platform Taobao\footnote{\url{https://www.taobao.com}}.
  It is commonly used to test the multi-turn response selection models \cite{Zhang2018ModelingMC,Yuan2019MultihopSN}.
  \item \textbf{Douban Corpus} \cite{Wu2017SequentialMN} is another popular response selection dataset, 
  which contains dyadic dialogs crawle from the Douban Group\footnote{\url{https://www.douban.com/group}}.
  It should be noted that, in the original test dataset, each conversation context may have multiple ground-truths,
  and we ignore these cases in this paper. 
  \item \textbf{Zh50w Corpus}\footnote{\url{https://github.com/yangjianxin1/GPT2-chitchat}} is a chinese open-domain dialog corpus.
  It is crawled from the Weibo social network platform,
  which has more casual conversations than Douban Corpus and E-Commerce Corpus.
  \item \textbf{LCCC Corpus} \cite{Wang2020ALC} is a large-scale cleaned chinese open-domain conversation dataset.
  The quality of LCCC Corpus is ensured by a rigorous data cleaning pipeline, 
  which is built based on a set of rules and a classifier.
  The size of the original LCCC Corpus is very huge, 
  and we randomly sample 2 million conversations in this paper.
\end{itemize}

For each corpus, we save all of the responses in train and test datasets into corresponding pre-constructed database,
which is used by corase-grained selection module.
The details of these datasets are shown in Table \ref{tab:1}.

\begin{table}[h]
  \resizebox{0.5\textwidth}{!}{
  \begin{tabular}{c|c|c|c|c}
  \hline
  \textbf{Datasets}   & \textbf{Train} & \textbf{Test} & \textbf{Retrieval Ratio} & \textbf{Database Size} \\ \hline
  \textbf{E-Commerce} & 1M            & 1,000          & 46.81\%      & 109,105             \\ \hline
  \textbf{Douban}     & 1M            & 667            & 54.57\%      & 442,280             \\ \hline
  \textbf{Zh50w}      & 1M            & 3,000          & 28.5\%       & 388,614             \\ \hline
  \textbf{LCCC}       & 4M            & 10,000         & 33.59\%      & 1,651,899           \\ \hline
  \end{tabular}
  }
  \caption{Data statistic of four popular datasets. 
  \textbf{Retrieval Ratio} is the proportion of samples that can be retrieved by sparse representation methods.
  \textbf{Database Size} is the number of the utterances saved in pre-constructed database.}
  \label{tab:1}
\end{table}

\subsection{Methods}

In this paper, three coarse-grained selection methods are measured:
(1) \textbf{BM25} \cite{Robertson2009ThePR}: 
following the previous works \cite{Karpukhin2020DensePR,Xiong2020ApproximateNN,Luan2020SparseDA}, 
we select BM25 sparse representation method,
which is widely used in real scenarios;
(2) \textbf{Dense} \cite{Karpukhin2020DensePR}: 
we select Dense as the dense representation method,
which use the PLM-based dual-encoder architecture to construct the coarse-grained selection module 
\cite{Karpukhin2020DensePR,Luan2020SparseDA};
(3) \textbf{DSHC}: our proposed deep semantic hashing based coarse-grained selection method.
More details are shown in Section \ref{sec:1}.

To implement BM25 method, Elasticsearch\footnote{\url{https://www.elastic.co}} is used in this paper, 
which is a very powerful search engine based on the Lucene library.
For Dense and DSHC methods, following the previous works \cite{Xiong2020ApproximateNN,Karpukhin2020DensePR}, 
FAISS\footnote{\url{https://github.com/facebookresearch/faiss}} \cite{JDH17,Karpukhin2020DensePR} toolkit is used in this paper.
Besides, GPU devices (GeForce GTX 1080 Ti) are used to accelerate the searching process.

\subsection{Evaluation Metrics}
To measure the performance of these coarse-grained selection modules in real scenarios,
we select four important evaluation metrics:

\textbf{Effectiveness}: Following previous work \cite{Xiong2020ApproximateNN,Karpukhin2020DensePR}, 
    Coverage@20/100 (Top-20/100) metric is used to evaluate 
    whether Top-20/100 retrieved candidates include the ground-truth response.
    However, during the testing, we find that this metric is not appropriate 
    to measure the effectiveness of the coarse-grained selection module.
    The reasons are as follows:
    The Top-20/100 metric only demonstrates whether only one ground-truth response can be retrieved.
    It cannot reflect the quality of all of the retrieved candidates.
    A good coarse-grained selection module should recalls candidates 
    that are all semantic coherent with the given conversation context, not only one candidate.
    Thus, in this paper, we propose a Correlation@20/100 (Correlation-20/100)
    as a more reliable metric to measure the effectiveness.
    Specifically, we leverage a state-of-the-art fine-grained selection module 
    \cite{Whang2019DomainAT} that fine-tunes on BERT model,
    to provide average correlation scores of the retrieved candidates.

\textbf{Search Time Cost}: 
  Search time cost is a core metric in a real application,
  which directly influences the interaction speed between chatbots and customers.
  In this paper, we records the average time cost (milliseconds)
  that the coarse-grained selection module searches the $bsz$ candidate responses for each conversation context in test dataset, 
  where $bsz=16$.

\textbf{Index Stoarge}: 
  For every coarse-grained selection module,
  it will construct a index that is calculated off-line to search the candidates.
  For sparse representation method, 
  the index is a inverted index storing a mapping from keywords to its locations in a candidate response.
  For dense representation method,
  the index is a huge matrix $M\in \mathbb{R}^{n\times d}$ that saves the dense vectors of all candidate responses,
  where $n$ is the number of the utterances in the pre-constructed database, and $d$ is the length of vectors.

\textbf{Human Evaluation}:
  In dialog system research community, human evaluation is the most reliable metric to measure the performance of dialog systems 
  \cite{Liu2016HowNT,Tao2018RUBERAU}.
  In this paper, for each corpus, three crowd-sourced annotators are employed to evaluate 
  the quality of generated responses for 200 randomly sampled conversation context.
  It should be noted that, the responses are generated by a whole retrieval-based chatbot,
  which consists of one coarse-grained selection module (BM25 or Dense or DSHC) and 
  a state-of-the-art fine-grained selection module \cite{Whang2019DomainAT}.
  During the evaluation, the annotators are requested to select a preferred response, or vote a tie
  from two responses that are generated by two retrieval-based chatbots.
  Besides, Cohen's kappa scores \cite{Cohen1960ACO} are used to 
  measure the intra-rater reliability.

\section{Comparison of Sparse and Dense Representation Methods}
In this section, we measure the performance of two kinds of coarse-grained selection module.
The experiment results are shown in Table \ref{tab:2} and Table \ref{tab:3}, 
and we can make the following conclusions:

\textbf{Effectiveness}: As shown in Table \ref{tab:3}, it can be observed that,
  dense representation method show the worse performance than BM25 method on Top-20/100 metrics.
  As described before, the Top-20/100 metrics are questionable to measure the quality of the retrieved candidates,
  because Top-20/100 metrics cannot consider the average coherence between the candidates and the given converesation context.
  As for the Correlation-20/100 metrics, it can be found that the dense representation significantly outperforms the BM25 method.
  For example, compared with BM25 method,
  dense representation method achieves average 19.89\% absolute improvement on Correlation-20 metric,
  which demonstrates that the candidates retrieved by dense representation method are more semantic coherent with the conversation context.

  \textbf{Index Storage}: Referring to the results in sixth columns in Table \ref{tab:3}, it can be observed that
  the dense representation method has more than 200 times the average index storage occupation 
  of the BM25 method.
  As shown in Figure \ref{img:3} (b), it can also be observed that, the index stoarge is even much bigger than the pre-constructed database.
  The index storage becomes too big to use in real scenarios as
  more and more utterances are saved in pre-constructed database.

  \textbf{Search Time Cost}: 
  Referring to the results in the seventh column in Table \ref{tab:3}, 
  although the computational complexity of Dense method is much bigger than BM25 method,
  Dense achieves the smaller searching time cost than BM25 method on E-Commerce Corpus and Douban Corpus,
  with the help of the parallel computing provided by GPU devices.
  However, if the size of the pre-constructed database becomes huge,
  the dense representation method still cost more time than BM25 method, for example, LCCC Corpus.

  \textbf{Human Evaluation}: As shown in Table \ref{tab:2}, it can be found that dense representation method brings more
  preferable responses compared with BM25 method on these four datasets,
  which indicates that rich semantic information captured by dense representation 
  does improve the response quality.

\begin{table}[H]
  \resizebox{0.5\textwidth}{!}{
  \begin{tabular}{c|c|c|c|c}
  \hline
  \textbf{Dense vs. BM25} & \textbf{Win} & \textbf{Loss} & \textbf{Tie} & \textbf{Kappa} \\ \hline
  \textbf{E-Commerce}     & \textbf{0.5917} &	0.2117&	0.1967&	0.7679	  \\ \hline
  \textbf{Douban}         & \textbf{0.4783} &	0.1883&	0.3333&	0.8240    \\ \hline
  \textbf{Zh50w}          & \textbf{0.5017} &	0.2683&	0.23  &	0.7143	  \\ \hline
  \textbf{LCCC}           & \textbf{0.5233} &	0.305 &	0.1717&	0.5558  \\ \hline
  \end{tabular}
  }
  \caption{Human evaluation of \textbf{Dense vs. BM25} on four datasets. 
  Very high Cohen's kappa scores prove the high consistency among the annotators.}
  \label{tab:2}
\end{table}

Compared with BM25 method, dense representation method could achieve better performance 
but cost more time and index storage occupation, which is unsatisfied in real scenarios.
In order to overcome these fatal weaknesses, in next sestion, 
we propose a novel deep semantic hashing based coarse-grained selection module, called DSHC.

\begin{table*}[h]
  \centering
    \resizebox{\textwidth}{!}{
      \subtable[Experiment results on E-Commerce Corpus.]{
      \begin{tabular}{c|c|c|c|c|c|c}
      \hline
      \textbf{Methods}    & \textbf{Top-20} & \textbf{Top-100} & \textbf{Correlation-20} & \textbf{Correlation-100} & \textbf{Index Storage} & \textbf{Search Time Cost (20/100)} \\ \hline
      \textbf{BM25}       & 0.025           & 0.055            & 0.615                   & 0.5122                   & \textbf{2.9 Mb}           & 89.5ms/129.4ms                       \\ \hline
      \textbf{Dense (gpu)} & \textbf{0.204}           & \textbf{0.413}            & \textbf{0.9537}                  & \textbf{0.9203}                   & 320 Mb           & \textbf{40.6ms/39.8ms}                  \\ \hline
      \end{tabular}
      }
    }
    \qquad
    \resizebox{\textwidth}{!}{
      \subtable[Experiment results on Douban Corpus.]{
      \begin{tabular}{c|c|c|c|c|c|c}
      \hline
      \textbf{Methods}    & \textbf{Top-20} & \textbf{Top-100} & \textbf{Correlation-20} & \textbf{Correlation-100} & \textbf{Index Storage} & \textbf{Search Time Cost (20/100)} \\ \hline
      \textbf{BM25}       &  	\textbf{0.063} &	0.096	 & 0.6957	& 0.6057&	\textbf{21.4 Mb}&	448.7ms/499.7ms   \\ \hline
      \textbf{Dense (gpu)} &   0.054	& \textbf{0.1049} & 	\textbf{0.9403} & 	\textbf{0.9067}	&1.3 Gb&	\textbf{200ms/177.1ms}    \\ \hline
      \end{tabular}
      }
    }
    \qquad
    \resizebox{\textwidth}{!}{
      \subtable[Experiment results on Zh50w Corpus.]{
      \begin{tabular}{c|c|c|c|c|c|c}
      \hline
      \textbf{Methods}    & \textbf{Top-20} & \textbf{Top-100} & \textbf{Correlation-20} & \textbf{Correlation-100} & \textbf{Index Storage} & \textbf{Search Time Cost (20/100)} \\ \hline
      \textbf{BM25}       & \textbf{0.0627}&	\textbf{0.1031}&	0.84&	0.7341	&\textbf{10.8 Mb}&	\textbf{91.5ms/122.8ms} \\ \hline
      \textbf{Dense (gpu)} & 0.044	&0.0824&	\textbf{0.9655}	&\textbf{0.9424}	&1.2 Gb	&122.4ms/128.3ms   \\ \hline
      \end{tabular}
      }
    }
    \qquad
    \resizebox{\textwidth}{!}{
      \subtable[Experiment results on LCCC Corpus.]{
      \begin{tabular}{c|c|c|c|c|c|c}
      \hline
      \textbf{Methods}    & \textbf{Top-20} & \textbf{Top-100} & \textbf{Correlation-20} & \textbf{Correlation-100} & \textbf{Index Storage} & \textbf{Search Time Cost (20/100)} \\ \hline
      \textbf{BM25}       & \textbf{0.0376}&	0.07&	0.8966	&0.8253	&\textbf{44 Mb}&	\textbf{190.1ms/247ms}   \\ \hline
      \textbf{Dense (gpu)} & 0.0351	&\textbf{0.0778}&	\textbf{0.9832}&	\textbf{0.9726}&	4.8 Gb	&458.6ms/572.2ms \\ \hline
      \end{tabular}
      }
    }
    \caption{The comparison between the BM25 method and Dense method. Dense method significantly outperforms BM25 method,
    but cost more time and index stoarge occupation.}
    \label{tab:3}
\end{table*}

\section{Deep Semantic Hashing Coarse-grained Selection Method (DSHC)} \label{sec:1}

\subsection{Methodology}

\subsubsection{The Architecture of DSHC}
\begin{figure}[h]     
  \center{\includegraphics[width=0.5\textwidth, height=8.5cm] {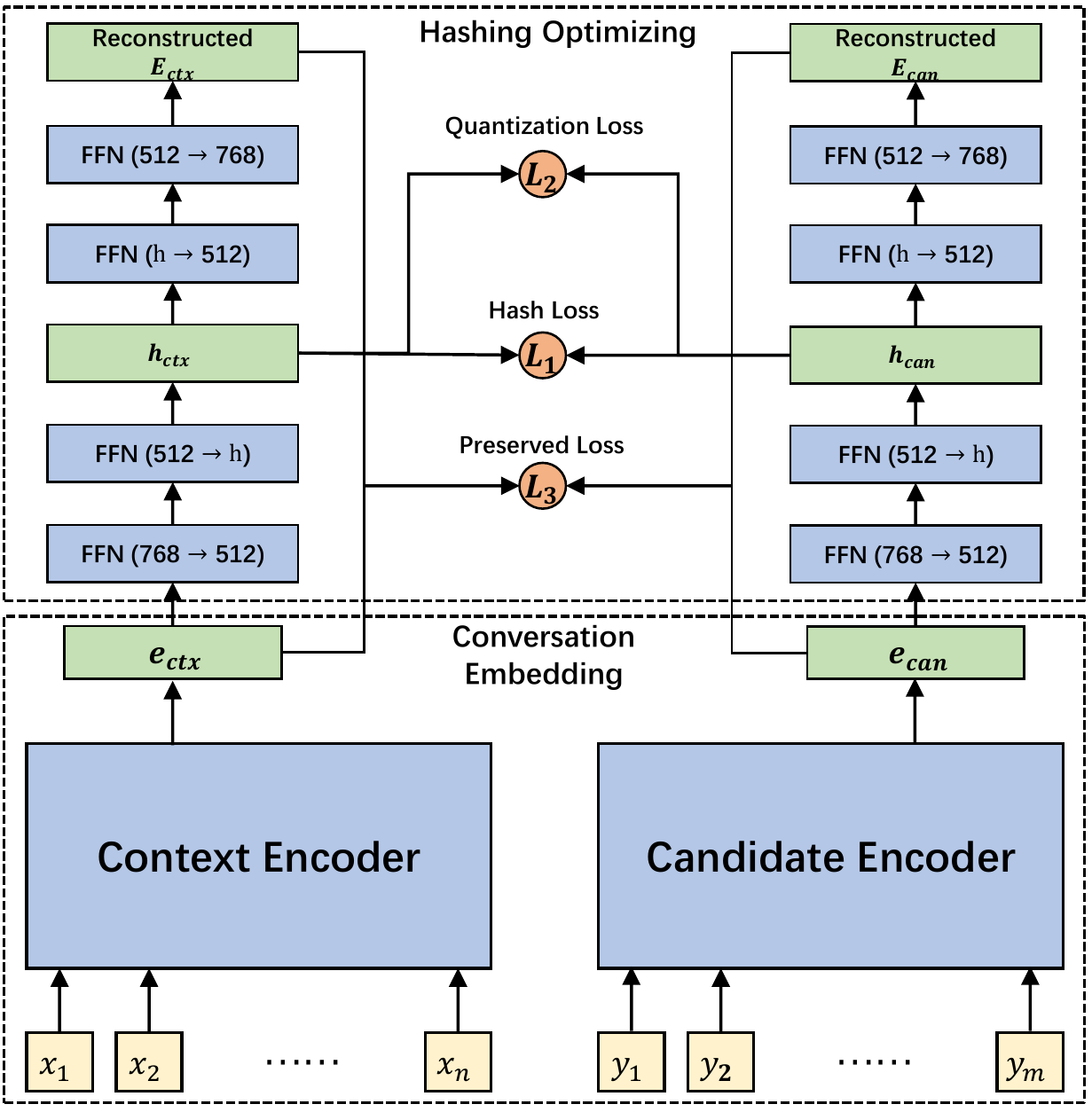}}        
  \caption{The overview of our proposed DSHC model for retrieval-based chatbots.
  DSHC model contains two parts: \textbf{conversation embedding} and \textbf{hashing optimizing}.}
  \label{img:2}
\end{figure}

The overview of our proposed DSHC model is shown in Figure \ref{img:2}, which contains two parts:
\textbf{conversation embedding} and \textbf{hashing optimizing}.

For \textbf{conversation embedding} part, we leverage a trained dense representation coarse-grained selection module.
Given a conversation context $\{x_i\}_{i=1}^n$ and a candidate response $\{y_i\}_{i=1}^m$,
where $n$ and $m$ is the number of the tokens, 
the conversation embedding separately encodes them into the dense embeddings $e_{ctx}$ and $e_{can}$.
The trained dense representation method ensures that the dense vector of an appropriate response $e_{can}$
is very similar to the dense vector of context embedding $e_{ctx}$, otherwise it is not.

For \textbf{hashing optimizing} part, DSHC model optimizes two 
deep autoencoders to generate the hash codes $h_{ctx}$ and $h_{can}$ for $e_{ctx}$ and $e_{can}$
by minimising the objective function that consists of three loss functions: 
\textbf{quantization loss}, \textbf{hash loss}, and \textbf{preserved loss}.
Specifically, the hashing optimizing part first encodes the dense embeddings into the
output vectors $o_{ctx}$, $o_{can}$:
\begin{equation}
  \begin{split}
    & o_{ctx} = {\rm Encoder}_{ctx}(e_{ctx}), o_{ctx}\in \mathbb{R}^{{\rm h}} \\
    & o_{can} = {\rm Encoder}_{can}(e_{can}), o_{can}\in \mathbb{R}^{{\rm h}} \\
    & h_{ctx} = sign(o_{ctx}), \{-1,1\}^{\rm h} \\
    & h_{can} = sign(o_{can}), \{-1,1\}^{\rm h} \\
  \end{split}
\end{equation},
where ${\rm h}$ is the hash code size.
During inference, $sign(\cdot)$ function is used to convert $o_{ctx}$ and $o_{can}$ 
into the hash codes $h_{ctx}$ and $h_{can}$.
Then, hashing optimizing part reconstructs the dense embeddings from $o_{ctx}$ and $o_{can}$:
\begin{equation}
  \begin{split}
    & E_{ctx} = {\rm Decoder}_{ctx}(o_{ctx}), E_{ctx}\in \mathbb{R}^{768} \\
    & E_{can} = {\rm Decoder}_{can}(o_{can}), E_{can}\in \mathbb{R}^{768} \\
  \end{split}
\end{equation}, where $E_{ctx}$ and $E_{can}$ are the reconstructed dense embeddings,
which assists to optimize the hash codes. 

\subsubsection{Objective Function}

Our proposed DSHC model aims to compressed the dense vectors $e_{ctx}$ and $e_{can}$ into
semantic similarity-preserving hash codes $h_{ctx}$ and $h_{can}$ that
can be efficiently computed in real scenarios.
Besides, the hash code of an appropriate response $h_{can}$ should be very similar to 
the hash code of the conversation context $h_{ctx}$, otherwise it is not.
In order to achieve this goal, we design three loss functions to optimize the hashing optimizing part:
(1) preserved loss; (2) hash loss; (3) quantization loss.

\textbf{Preserved loss}:
To preserve rich semantic information in dense vectors into hash codes,
the reconstructed dense embeddings $E_{ctx}$ and $E_{can}$ should be similar to
$e_{ctx}$ and $e_{can}$.
Thus, we design the preserved loss to measure the difference
between $e_{ctx}$ and $E_{ctx}$, and between $e_{can}$ and $E_{cab}$, 
which are the L2 norm (Euclidean norm) losses:
\begin{equation}
  \begin{split}
    & \mathcal{L}_{p}=\|e_{ctx}-E_{ctx}\|_{2}+\|e_{can}-E_{can}\|_{2} \\
  \end{split}
\end{equation}

\textbf{Hash loss}:
Although preserved loss ensures that the $o_{ctx}$ and $o_{can}$ contains the rich
semantic information in $e_{ctx}$ and $e_{can}$,
there is still no way to measure the similarity between the conversation context hash codes and candidate hash codes.
In order to ensure that hash codes could preserve the semantic similarity between
the conversation context and the candidate response, the hash loss is designed.
For hash codes in Hamming space, 
if the similarity $S(o_{ctx}, o_{can})=1$,
i.e., the candidate is appropriate to the context,
the Hamming distance $\|o_{ctx}, o_{can}\|_{H}=0.5({\rm h}-o_{ctx}^To_{can})$
between $o_{ctx}$ and $o_{can}$ should be equal to 0,
which indicates that $o_{ctx}^To_{can}$ should be equal to ${\rm h}$,
where ${\rm h}$ is the dimension of the hash codes;
if the similarity $S(o_{ctx}, o_{can})=0$,
i.e., the candidate is inappropriate to the context,
the Hamming distance $\|o_{ctx}, o_{can}\|_{H}$ should be equal to $\frac{{\rm h}}{2}$,
which indicates that $o_{ctx}^To_{can}$ should be equal to 0.
Therefore, the hash loss is designed as following:
\begin{equation}
  \begin{split}
    & \mathcal{L}_{h}=\|o_{ctx}^To_{can}-{\rm h}S(o_{ctx},o_{can})\|_{2} \\
    & \text{ s.t. } \quad S(o_{ctx},o_{can})\in \{0, 1\} \\
  \end{split}
\end{equation}

\textbf{Quantization loss}:
So far, preserved loss and hash loss ensure that $o_{ctx}$ and $o_{can}$ 
preseve the semantic information and the similarity between them.
However, during inference, the hash codes $h_{ctx}$ and $h_{can}$ are used to search the candidates,
which are roughly converted by using $sign(\cdot)$ function.
In order to narrow the gap between $h_{ctx}$ and $o_{ctx}$, and $h_{can}$ and $o_{can}$,
the quantization loss \cite{Wang2018ASO} is used to ensure that
each element of $o_{ctx}$ and $o_{can}$ can be close to ``+1'' or ``-1'':
\begin{equation}
  \begin{split}
    & \mathcal{L}_{q}=\|h_{ctx}-o_{ctx}\|_{2} + \|h_{can}-o_{can}\|_{2} \\
  \end{split}
\end{equation}

Finally, the overall objective function is obtained as follows:
\begin{equation}
  \begin{split}
    & \mathcal{L}=\mathcal{L}_{p} + \mathcal{L}_{h} + \gamma_t\cdot \mathcal{L}_{q} \\
    & \text{s.t.} \quad \gamma_t=\gamma_{min} + \frac{\gamma_{max}-\gamma_{min}}{T}\cdot t \\
  \end{split}
\end{equation},
where $\gamma_t$ is a hyperparameter that dynamically balances the processing of optimizing hash loss and quantization loss.
In this paper, $\gamma_{min}=1e-4$, $\gamma_{max}=1e-1$.
$T$ is the number of the mini-batch in one epoch, and $t\in \{0,1,2...,T-1\}$ is the current running step.

\subsection{Overall Comparison}
In this section, we carefully compare three coarse-grained selection methods:
(1) BM25; (2) Dense; (3) our proposed DSHC model.

\textbf{Effectiveness}:
As shown in Table \ref{tab:4}, it can be observed that,
our proposed DSHC model significantly outperforms the BM25 method.
Besides, the performance of DSHC model is very close to the Dense representation method,
which indicates that our proposed DSHC model effectively preserves the rich semantic information
and the similarity information between the conversation context and candidate response.
For example, there is only 2.6\% absolute average decline on correlation-20 metric for DSHC-512 model.
In view of that compressed binary hash codes lost lots of information,
the results are pretty good.

\textbf{Index Storage}:
Furthermore, as shown in sixth column in Table \ref{tab:4}, 
it can also be found that, the index storage occupation of our proposed DSHC model is 
much smaller than the Dense method, 
even smaller than BM25 method if the dimension of the hash codes is 128.

\textbf{Search Time Cost}:
Moreover, although the computational complexity of computing hamming distance is worse than BM25,
with the help of very high computational efficiencies of hash codes, and parallel computing provided by GPU devices,
our proposed DSHC model still achieves the smallest search time cost i.e. the fastest search speed, than BM25 methods.
For example, DSHC-128 model is nearly 15x faster than the widely used BM25 method.

\textbf{Human Evaluation}:
Finally, we also conduct the human evaluation to measure the performane more accurately.
As shown in Table \ref{tab:5} (a), it can be found that, 
the performance of DSHC and Dense methods are very close.
Quite surprisingly, our proposed DSHC model is even better than Dense method on LCCC corpus.
Beides, from Table \ref{tab:5} (b), it can also be found that, 
DSHC model significantly outperforms the widely used BM25 method,
because DSHC model wins most of the time.
The very high Cohen's kappa scores demonstrate that the decision of the annotators are highly consistent.


\begin{table*}[ht]
  \centering
    \resizebox{\textwidth}{!}{
      \subtable[Experiment results on E-Commerce Corpus.]{
      \begin{tabular}{c|c|c|c|c|c|c}
      \hline
      \textbf{Methods}    & \textbf{Top-20} & \textbf{Top-100} & \textbf{Correlation-20} & \textbf{Correlation-100} & \textbf{Index Storage} & \textbf{Search Time Cost (20/100)} \\ \hline
      \textbf{BM25}       & 0.025           & 0.055            & 0.615                   & 0.5122                   & 2.9 Mb           & 89.5ms/129.4ms                       \\ \hline
      \textbf{Dense (gpu)} & 0.204           & \textbf{0.413}            & \textbf{0.9537}                  & \textbf{0.9203}                   & 320 Mb           & 389.3ms/401.5ms                       \\ \hline
      \textbf{DSHC-128 (gpu)} & 0.185	&0.366	&0.9252	&0.8808	&\textbf{1.7 Mb}&	\textbf{4ms/6.3ms}   \\ \hline
      \textbf{DSHC-512 (gpu)} & \textbf{0.214}	&0.382&	0.944	&0.9065	&6.7 Mb	&9.3ms/18.7ms \\ \hline
      \end{tabular}
      }
    }
    \qquad
    \resizebox{\textwidth}{!}{
      \subtable[Experiment results on Douban Corpus.]{
      \begin{tabular}{c|c|c|c|c|c|c}
      \hline
      \textbf{Methods}    & \textbf{Top-20} & \textbf{Top-100} & \textbf{Correlation-20} & \textbf{Correlation-100} & \textbf{Index Storage} & \textbf{Search Time Cost (20/100)} \\ \hline
      \textbf{BM25}       & \textbf{0.063}&	0.096&	0.6957&	0.6057&	21.4 Mb	&448.7ms/499.7ms            \\ \hline
      \textbf{Dense (gpu)} & 0.054	&\textbf{0.1049}	&\textbf{0.9403}&	\textbf{0.9067}&	1.3 Gb&	200ms/177.1ms       \\ \hline
      \textbf{DSHC-128 (gpu)} & 0.012	&0.0465	&0.8375	&0.8016&	\textbf{6.8 Mb}&	\textbf{20.9ms/19.6ms} \\ \hline
      \textbf{DSHC-512 (gpu)} & 0.0225&	0.066	&0.8838	&0.8474&	27 Mb	&52.3ms/45.2ms
      \\ \hline
      \end{tabular}
      }
    }
    \qquad
    \resizebox{\textwidth}{!}{
      \subtable[Experiment results on Zh50w Corpus.]{
      \begin{tabular}{c|c|c|c|c|c|c}
      \hline
      \textbf{Methods}    & \textbf{Top-20} & \textbf{Top-100} & \textbf{Correlation-20} & \textbf{Correlation-100} & \textbf{Index Storage} & \textbf{Search Time Cost (20/100)} \\ \hline
      \textbf{BM25}       & \textbf{0.0627}&	\textbf{0.1031}	&0.84	&0.7341	&10.8 Mb&	91.5ms/122.8ms  \\ \hline
      \textbf{Dense (gpu)} & 0.044&	0.0824&	\textbf{0.9655}&	\textbf{0.9424}&	1.2 Gb&	122.4ms/128.3ms  \\ \hline
      \textbf{DSHC-128 (gpu)} & 0.027	&0.0724	&0.9108&	0.8835&	\textbf{6.0 Mb}&	\textbf{13.7ms/19.2ms}   \\ \hline
      \textbf{DSHC-512 (gpu)} & 0.0377	&0.0934&	0.944&	0.9223&	24 Mb	&23.5s/28ms  \\ \hline
      \end{tabular}
      }
    }
    \qquad
    \resizebox{\textwidth}{!}{
      \subtable[Experiment results on LCCC Corpus.]{
      \begin{tabular}{c|c|c|c|c|c|c}
      \hline
      \textbf{Methods}    & \textbf{Top-20} & \textbf{Top-100} & \textbf{Correlation-20} & \textbf{Correlation-100} & \textbf{Index Storage} & \textbf{Search Time Cost (20/100)} \\ \hline
      \textbf{BM25}       & \textbf{0.0376}	&0.07	&0.8966	&0.8253&	44 Mb&	190.1ms/247ms                  \\ \hline
      \textbf{Dense (gpu)} & 0.0351	&\textbf{0.0778}	&\textbf{0.9832}	&\textbf{0.9726}&	4.8 Gb&	458.6ms/572.2ms    \\ \hline
      \textbf{DSHC-128 (gpu)} & 0.014	&0.0348&	0.9369&	0.9187&	\textbf{26 Mb}	&\textbf{20.4ms/24.4ms}    \\ \hline
      \textbf{DSHC-512 (gpu)} & 0.0204	&0.0494&	0.9663&	0.9526	&101 Mb	&76.4ms/94ms  \\ \hline
      \end{tabular}
      }
    }
    \caption{Parameters 128 and 512 are the dimension of the hash codes ${\rm h}$ in our proposed DSHC model.}
    \label{tab:4}
\end{table*}

\begin{table}[H]
  \resizebox{0.5\textwidth}{!}{
    \subtable[Human Evaluation of \textbf{Dense vs. DSHC}.]{
    \begin{tabular}{c|c|c|c|c}
    \hline
    \textbf{Dense vs. DSHC} & \textbf{Win} & \textbf{Loss} & \textbf{Tie} & \textbf{Kappa} \\ \hline
    \textbf{E-Commerce}     & 0.3133	&0.2683& \textbf{0.4183}&	0.7025	  \\ \hline
    \textbf{Douban}         & 0.375	&0.2217	&\textbf{0.4033}&	0.8251    \\ \hline
    \textbf{Zh50w}          & \textbf{0.395}	&0.2833	&0.3217&	0.6679	  \\ \hline
    \textbf{LCCC}           & 0.3283&	\textbf{0.3733} &	0.2983&	0.7716  \\ \hline
    \end{tabular}
    }
  }
  \qquad
  \resizebox{0.5\textwidth}{!}{
    \subtable[Human Evaluation of \textbf{DSHC vs. BM25}.]{
    \begin{tabular}{c|c|c|c|c}
    \hline
    \textbf{DSHC vs. BM25} & \textbf{Win} & \textbf{Loss} & \textbf{Tie} & \textbf{Kappa} \\ \hline
    \textbf{E-Commerce}     & \textbf{0.6017}	&0.1933	&0.205&	0.6733	  \\ \hline
    \textbf{Douban}         & \textbf{0.4767}	&0.2783	&0.245&	0.8506   \\ \hline
    \textbf{Zh50w}          & \textbf{0.4733}	&0.335&	0.1917&	0.727	  \\ \hline
    \textbf{LCCC}           & \textbf{0.5317}	&0.27	&0.1983&	0.7115  \\ \hline
    \end{tabular}
    }
  }
  \caption{Human evaluation on four datasets. 
  Very high Cohen's kappa scores prove the high consistency among the annotators.}
  \label{tab:5}
\end{table}

\section{Case Study}
Due to the page limitation, cases are shown in Table \ref{tab:8} in \textit{Appendix}.
Refering to these cases, it can be found that,
the retrieval-based chatbots that use the dense representation and our proposed DSHC methods
provide more semantic coherent responses to the given conversation context than BM25 method.
Besides, the responses given by dense representation method and DSHC method are both very appropriate,
which proves the effectiveness of our proposed DSHC model.

\section{Conclusion}
In this paper, we first systematically compare the dense and sparse representation method
in retrieval-based chatbot from four important aspects: (1) effectiveness; (2) search time cost; (3) index stoarge; (4) human evaluation.
Extensive experiment results demonstrate that dense representation method could achieve better performance 
at the expense of more time cost and higher storage occupation,
In order to overcome these fatal weaknesses, we propose a deep semantic hashing based corase-grained (DSHC) selection method.
Extensive experiment results prove the effectiveness and the efficiency of DSHC model.

\bibliography{anthology,acl2020}
\bibliographystyle{acl_natbib}

\appendix

\section{Appendices}

\subsection{Computational Complexity of Search Operation}

The computational complexity of three coarse-grained selection methods are shown in Table \ref{tab:6}.
Although BM25 method achieves the best computational complexity by using well designed data structure, 
such as inverted index and skiplist,
it cannot be accelerated by using GPU devices.
In real scenarios, with the help of the parallel computing provided by GPU devices,
DSHC method could achieve much faster searching speed.

It should be noted that, 
lots of works have been proposed to optimize the computational complexity of 
computing the dot production and Hamming distance, such as product quantizer and inverted index,
and the computational complexities of Dense and DSHC methods shown in Table \ref{tab:6} are the worst cases. 
In this paper, we dont't consider to leverage these techniques to search candidates in Dense and DSHC methods.
Brute-force search i.e. linear scan is used to find the Top-K (20/100) candidates in coarse-grained selection module,
which directly scans all of the utterances in the pre-constructed database.

\begin{table}[h]
  \resizebox{0.5\textwidth}{!}{
    \begin{tabular}{cc}
    \hline
    \textbf{Coarse-grained Selection} & \textbf{Computational Complexity} \\ \hline
    \textbf{BM25 (Inverted index)}    & $O(\log n)$                       \\ 
    \tabincell{c}{\textbf{Dense (Dot production)}\\ \textit{brute-force}}   & $O(d\cdot n)$      \\ 
    \tabincell{c}{\textbf{DSHC (Hamming distance)}\\ \textit{brute-force}}  & $O(n)$             \\ \hline
    \end{tabular}
  }
  \caption{The computational complexity of the different coarse-grained selection methods.
  $n$ is the number of the utterances in the pre-constructed database.
  $d$ is the dimension of the dense vectors.}
  \label{tab:6}
\end{table}

\subsection{Hyperparameters Analysis}

In this section, we analyze the hyperparameter ${\rm h}$ i.e. 
the dimension of the hash codes, in our proposed DSHC model.
For our proposed DSHC model, we separately test the 16,32,48,64,128,256,512,1024 dimensions of the hash codes.
The results are shown in Table \ref{tab:7}.

\begin{table}[h]
  \resizebox{0.5\textwidth}{!}{
    \subtable[Hyperparameters in E-Commerce Corpus.]{
    \begin{tabular}{|c|c|c|c|c|}
    \hline
    \textbf{Methods}   & \textbf{Correlation-20} & \textbf{Correlation-100} & \textbf{Storage} & \textbf{Time Cost} \\ \hline
    \textbf{BM25}      & 0.615 &	0.5122&	8.8 Mb&	89.5ms/129.4ms \\ \hline \hline
    \textbf{DSHC-16}   & 0.6819&	0.6284&	\textbf{214 Kb} &	2.6ms/5.1ms  \\ \hline
    \textbf{DSHC-32}   & 0.8271&	0.7714&	427 Kb&	\textbf{1.7ms/3.1ms}   \\ \hline
    \textbf{DSHC-48}   & 0.8737&	0.8136&	640 Kb&	6.4ms/8.1ms   \\ \hline
    \textbf{DSHC-64}   & 0.8942&	0.8364&	853 Kb&	1.8ms/4.4ms  \\ \hline
    \textbf{DSHC-128}  & 0.9278&	0.8837&	1.7 Mb&	2.3ms/4.4ms   \\ \hline
    \textbf{DSHC-256}  & 0.9376&	0.8976&	3.4 Mb&	3.2ms/10.6ms  \\ \hline
    \textbf{DSHC-512}  & 0.944 &	0.9065&	6.7 Mb&	9.2ms/16.4ms   \\ \hline
    \textbf{DSHC-1024} & \textbf{0.9473}&	\textbf{0.9134}&	14 Mb &	19.4ms/18.4ms   \\ \hline \hline
    \textbf{Dense}     & 0.9537&	0.9203&	320 Mb&	389.3ms/401.5ms \\ \hline
    \end{tabular}
    }
  }
  \qquad
  \resizebox{0.5\textwidth}{!}{
    \subtable[Hyperparameters in Douban Corpus.]{
    \begin{tabular}{|c|c|c|c|c|}
    \hline
    \textbf{Methods}   & \textbf{Correlation-20} & \textbf{Correlation-100} & \textbf{Storage} & \textbf{Time Cost} \\ \hline
    \textbf{BM25}      & 0.6957 &	0.6057&	21.4 Mb&	448.7ms/499.7ms \\ \hline \hline
    \textbf{DSHC-16}   & &&&  \\ \hline
    \textbf{DSHC-32}   & &&&   \\ \hline
    \textbf{DSHC-48}   & &&&   \\ \hline
    \textbf{DSHC-64}   & &&&  \\ \hline
    \textbf{DSHC-128}  & 0.8375	&0.8016&	6.8 Mb&	20.9ms/19.6ms   \\ \hline
    \textbf{DSHC-256}  & &&&  \\ \hline
    \textbf{DSHC-512}  & 0.8838&	0.8474&	27 Mb&	52.3ms/45.2ms   \\ \hline
    \textbf{DSHC-1024} & &&&   \\ \hline \hline
    \textbf{Dense}     & 0.9403&	0.9067&	1.3 Gb&	200ms/177.1ms \\ \hline
    \end{tabular}
    }
  }
  \qquad
  \resizebox{0.5\textwidth}{!}{
    \subtable[Hyperparameters in Zh50w Corpus.]{
    \begin{tabular}{|c|c|c|c|c|}
    \hline
    \textbf{Methods}   & \textbf{Correlation-20} & \textbf{Correlation-100} & \textbf{Storage} & \textbf{Time Cost} \\ \hline
    \textbf{BM25}      & 0.84	&0.7341&	10.8 Mb&	91.5ms/122.8ms \\ \hline \hline
    \textbf{DSHC-16}   & 0.6703&	0.6431&	\textbf{760 Kb}&	9.3ms/16.3ms  \\ \hline
    \textbf{DSHC-32}   & 0.7912&	0.7557&	1.5 Mb&	\textbf{5ms/6.5ms}   \\ \hline
    \textbf{DSHC-48}   & 0.8428&	0.8101&	2.3 Mb&	19.3ms/20.7ms   \\ \hline
    \textbf{DSHC-64}   & 0.8685&	0.836&	3.0 Mb&	5.1ms/6.3ms  \\ \hline
    \textbf{DSHC-128}  & 0.9108&	0.8835&	6.0 Mb&	9.4ms/17.4ms   \\ \hline
    \textbf{DSHC-256}  & 0.9353&	0.9105&	12 Mb &	16ms/13.3ms  \\ \hline
    \textbf{DSHC-512}  & 0.944&	0.9223&	24 Mb&	23.5ms/28ms   \\ \hline
    \textbf{DSHC-1024} & \textbf{0.9546}&	\textbf{0.9336}&	48 Mb&	50.2ms/64.7ms   \\ \hline \hline
    \textbf{Dense}     & 0.9655&	0.9424&	1.2 Gb&	122.4ms/128.3ms \\ \hline
    \end{tabular}
    }
  }
  \qquad
  \resizebox{0.5\textwidth}{!}{
    \subtable[Hyperparameters in LCCC Corpus.]{
    \begin{tabular}{|c|c|c|c|c|}
    \hline
    \textbf{Methods}   & \textbf{Correlation-20} & \textbf{Correlation-100} & \textbf{Storage} & \textbf{Time Cost} \\ \hline
    \textbf{BM25}      & 0.8966&	0.8253&	44 Mb&	190.1ms/247ms \\ \hline \hline
    \textbf{DSHC-16}   & &&&  \\ \hline
    \textbf{DSHC-32}   & &&&   \\ \hline
    \textbf{DSHC-48}   & &&&   \\ \hline
    \textbf{DSHC-64}   & &&&  \\ \hline
    \textbf{DSHC-128}  & 0.9369&	0.9187&	26 Mb&	20.4ms/24.4ms   \\ \hline
    \textbf{DSHC-256}  & &&&  \\ \hline
    \textbf{DSHC-512}  & 0.9663&	0.9526&	101 Mb&	76.4ms/94ms   \\ \hline
    \textbf{DSHC-1024} & &&&   \\ \hline \hline
    \textbf{Dense}     & 0.9832&	0.9726&	4.8 Gb&	458.6ms/572.2ms \\ \hline
    \end{tabular}
    }
  }
  \caption{The hyperparameters analysis in four datasets.}
  \label{tab:7}
\end{table}

\subsection{Case Study}

The whole cases in four datasets are shown in Table \ref{tab:8}, 
and all of the utterances are translated from Chinese to English.
More cases can be found in this page\footnote{\url{https://github.com/gmftbyGMFTBY/HashRetrieval}}.

\begin{table*}[h]
  \resizebox{\textwidth}{!}{
    \subtable[Cases in E-Commerce Corpus.]{
      \begin{tabular}{c|c|c|c|c}
      \hline
      \textbf{Conversation Context} & \textbf{Ground-truth} & \textbf{Dense} & \textbf{DSHC} & \textbf{BM25} \\ \hline
      \tabincell{c}{What about this delivery?\\ It's the slowest delivery ever} &  Hello?  & \tabincell{c}{I'm sorry that all the EMS items to \\Guangdong are a little slow recently.}  & \tabincell{c}{Hello, I'm sorry that the courier is off work here,\\ now I will help you leave a message to urge the courier.} & \tabincell{c}{This is the slowest estimate.}              \\ \hline
      \tabincell{c}{If I buy it today, when will it be shipped? \\ 35 days after purchase \\ It’s so late.}& Yes & \tabincell{c}{I'm really sorry\\there are so many orders before new year.}& \tabincell{c}{Shipped 35 days after purchase} & When did you buy it? \\ \hline
      \tabincell{c}{Why didn't you send it to me?} & \tabincell{c}{The dried durian has just arrived \\We will send it to you today or tomorrow.} & Already sent & \tabincell{c}{I'm really sorry \\ We are temporarily out of stock \\it is estimated that the goods \\wont be sent until the day after tomorrow} & \tabincell{c}{It's out of the warehouse\\why are you returning it?}\\ \hline
      \tabincell{c}{I will change the price for you. \\How long can the goods be delivered to Xinjiang?} & \tabincell{c}{We send goods from Jiaxing, Zhejiang. \\ Generally, it will cost 4-6 days after delivery.\\ Please wait patiently.} & About 23 days & 4,5 days after delivery & \tabincell{c}{I will change the price for you.} \\ \hline
      \tabincell{c}{After you buy it, please send it as soon as possible. \\ Ok, please check the delivery address. \\ It's Right}& \tabincell{c}{Okay, we will send it to you\\ as soon as possible.}& \tabincell{c}{Okay, we will send it to you as soon as possible.} & \tabincell{c}{Okay, we will send it to you as soon as possible.} & \tabincell{c}{Ok, please check the delivery address.} \\ \hline
      \tabincell{c}{Why did the laundry soap I bought \\ arrived in Changsha \\ and transferred to other places?} & \tabincell{c}{Hello, we will contact the courier to \\help you verify the situation.} & \tabincell{c}{Hello, we will help you contact the courier.} & \tabincell{c}{Hello, there should be a problem with the transit, \\please contact the courier.} & \tabincell{c}{It's impossible.} \\ \hline
      \tabincell{c}{What's the quality of walnuts?} & \tabincell{c}{The quality is very good.} & Pretty good. & Very good, very creamy. & \tabincell{c}{I'm really sorry for the large order quantity\\ I have kept you waiting}\\ \hline
      \tabincell{c}{Is there any discount on this product recently?\\ There will be a discount \\on the 16th of this month.\\ You can focus it. \\ Okay} & Yes. & \tabincell{c}{Please pay attention to this discount activity.} & \tabincell{c}{Please pay attention to this discount activity.} & \tabincell{c}{There will be a discount on the 16th of this month.} \\ \hline
      \tabincell{c}{Please send me some snacks to taste} & Okay, lol & Ok & Okay & You are a foodie lol \\ \hline
      \tabincell{c}{Which kind of courier company \\is used to deliver the goods?\\ We use ``Tian Tian express" by default\\ Please use EMS for delivery\\ Okay\\ I have already paid} & Okay & \tabincell{c}{Now, please check the delivery address\\We can send it to you today}& \tabincell{c}{Now, please check the delivery address\\We can send it to you today}& \tabincell{c}{We send goods by EMS} \\ \hline
      \end{tabular}
    }
  }
  \resizebox{\textwidth}{!}{
    \subtable[Cases in Douban Corpus.]{
      \begin{tabular}{c|c|c|c|c}
      \hline
      \textbf{Conversation Context} & \textbf{Ground-truth} & \textbf{Dense} & \textbf{DSHC} & \textbf{BM25} \\ \hline
      \tabincell{c}{I would like to ask where there\\ are custom-made clothes in Zhengzhou. \\ The leather market opposite the Chengdong Road Art Museum\\ makes custom-made leather goods. \\ Wow, I live nearby and I don’t know \\ Hey, we live so close} & How close is it? & Where do you live? & Lol, it's really close & I'm from Zhengzhou \\ \hline
      \tabincell{c}{Every day I shit smoothly, but I still get hemorrhoids\\ So do I. \\ Take good care of stomach} & \tabincell{c}{I also have a bad stomach\\so I don't eat chili recently} & \tabincell{c}{You have to take care of your stomach.} & \tabincell{c}{I have been drinking red bean and barley soup for a month,\\ and I feel better.} & So do I \\ \hline
      \tabincell{c}{It’s said that many companies don’t have year-end awards.\\ Are you talking about my company?\\ I didn’t mean it} & No, you did it on purpose. & \tabincell{c}{I mean mine, not yours, don't be angry} & \tabincell{c}{Alas, I’m mad at talking about benefits\\Our year-end bonus is too small} & You are insatiable \\ \hline
      \tabincell{c}{Lol, my sister has passed the exam\\ Which school does your sister go to?} & \tabincell{c}{Dalian University of Technology\\hope not far from you} & Ludong University & Ludong University & \tabincell{c}{My sister went to school in 2009} \\ \hline
      \tabincell{c}{I think hairy crabs are the most delicious.\\ What is it? I have never heard of it.\\ You must have eaten it.} & \tabincell{c}{Maybe, I didn’t recognize it when I ate} & \tabincell{c}{I ate once a long time ago\\and never eaten again} & \tabincell{c}{I have no impression at all\\either it’s too long or the taste is not good enough} & Hairy crabs in the fish market \\ \hline
      \tabincell{c}{After reading your post, I went to sign up.\\ Did you sign up successfully?\\ Yes \\What position did you sign up for?} & \tabincell{c}{I signed up for a national tax civil servant\\but the location is uncertain\\ how about you?} & \tabincell{c}{I have already signed up in the first half of the year\\ but I haven't started to sign up yet} & \tabincell{c}{I signed up for an oilfield civil service institution} & Have you signed up \\ \hline
      \end{tabular}
    }
  }
  \qquad
  \resizebox{\textwidth}{!}{
    \subtable[Cases in Zh50w Corpus.]{
      \begin{tabular}{c|c|c|c|c}
      \hline
      \textbf{Conversation Context} & \textbf{Ground-truth} & \textbf{Dense} & \textbf{DSHC} & \textbf{BM25} \\ \hline
      \tabincell{c}{Health is very important, \\I exercise as much as possible. \\ My leg hurts and I can't exercise recently. \\Your leg injury has not recovered, so take care of it slowly!\\ It has been almost three months.\\ Take good care and don't leave any sequelae!}&Thank you for your concern! & \tabincell{c}{Hmm, I recently went to\\ the hospital for a review.}&\tabincell{c}{Well, I have been practicing \\interval exercise recently.} & \tabincell{c}{Well, a natural birth will \\help you recover quickly}\\ \hline
      \tabincell{c}{What is the eighth picture?\\
      I haven't seen it in the north of China. \\
      The flower is golden triangle plum with white edges.} & 
      Thank you very much &
      Thanks! It's so beautiful. & 
      Thanks! It's so beautiful. & 
      \tabincell{c}{I know. Where is this flower} \\ \hline
      \tabincell{c}{Good morning!\\Good morning, thank you\\ Lol, Why are you so polite?} & 
      \tabincell{c}{It's not polite. \\I want to respect you} & 
      \tabincell{c}{I'm just kidding. \\I don't use twitter very often. \\Let's have a chat at WeChat.} & 
      \tabincell{c}{Lol, It's not polite. Good morning} & 
      What do you want to do \\ \hline
      \tabincell{c}{Your glasses look good!\\Oh? How about me} & 
      You look as good as glasses & 
      You look better than glasses! & 
      You look better than glasses! & 
      Your glasses are very artistic \\ \hline
      \tabincell{c}{Sister, why are you still up?\\ I go to bed late every day} &
      \tabincell{c}{Yes, many of your Weibo posts are posted early in the morning} & 
      I have to sleep & 
      \tabincell{c}{Insomnia is painful, sister. \\
      I didn’t expect you to sleep so late. \\
      if I knew I would chat with you} & 
      Good morning sister! \\ \hline
      \tabincell{c}{My friend sent me some mytilus edulis.\\
      Does anyone know how to cook it?\\
      Mytilus edulis, also known as oyster, \\
      can be cooked with soup, porridge, braised mushrooms.} & 
      Thank you very much & 
      Ok, thank you. I'll try it & 
      Okay, thank you for sharing & 
      \tabincell{c}{Cut it into thin slices\\ and put it in the soup} \\ \hline
      \end{tabular}
    }
  }
  \qquad
  \resizebox{\textwidth}{!}{
    \subtable[Cases in LCCC Corpus.]{
      \begin{tabular}{c|c|c|c|c}
      \hline
      \textbf{Conversation Context} & \textbf{Ground-truth} & \textbf{Dense} & \textbf{DSHC} & \textbf{BM25} \\ \hline
      \tabincell{c}{I want to repost this video\\ god please save me, he is so handsome} &
      \tabincell{c}{You are super-looking too!\\ I dare not look up when I see you} &
      \tabincell{c}{Yeah, I want to share him with the world.} & 
      \tabincell{c}{Yes, he is so good-looking, and I want to cry} &
      \tabincell{c}{Help me! I'm so wronged} \\ \hline
      \tabincell{c}{I feel exhausted to watch the show now, \\
      I recommend you a movie "Fruit Hard Candies", \\it's pretty good} & 
      I have seen it before &
      \tabincell{c}{I've seen it, very good and sweet} & 
      \tabincell{c}{I've seen it, by the way,\\ why do you recommend this one?} &
      I'm exhausted \\ \hline
      \tabincell{c}{I want to ask, where is this?} &
      Shanghai BMW Carnival &
      \tabincell{c}{In the underground plaza behind the small park} &
      \tabincell{c}{Opposite the gate of the South District \\of Xi'an University of Foreign Affairs} &
      I want to ask, where is this? \\ \hline
      \tabincell{c}{I also have a psychological shadow. \\
      I wont try this food for the second time.\\
      So let's eat something else!} &
      \tabincell{c}{Okay, let's eat other delicious food} &
      \tabincell{c}{What are we going to eat?} &
      \tabincell{c}{What are you goning to eat? \\Give me some suggestions} &
      I wont try this food for the second time \\ \hline
      \tabincell{c}{I have also been to Qinglong Town, \\
      the newly developed tourism classic this year, \\
      and it has not yet started to charge} &
      \tabincell{c}{This is a newly developed scenic spot\\ and there is no charge yet}&
      How can it be free? &
      \tabincell{c}{Free tourist attractions lose their value} &
      My charge is expensive \\ \hline\
      \tabincell{c}{Is Thai food delicious? \\
      Yummy! But you need to choose seriously, \\or you'll get very bad food.} &
      \tabincell{c}{Can you share your chosen restaurants?} & 
      \tabincell{c}{Lol thanks, I saw the restaurant\\ that you shared last night,\\ and I will try it tomorrow} &
      \tabincell{c}{I like your travel notes very much. \\ 
      As a foodie, I tasted all \\the delicious food you introduced.\\ 
      It was really delicious} &
      \tabincell{c}{I don't need your bad food} \\ \hline
      \end{tabular}
    }
  }
  \caption{Utterances are already translated from Chinese to English.}
  \label{tab:8}
\end{table*}

\subsection{Storage Occupation Visualization}
\begin{figure}[h]
  \centering
  \subfigure[BM25 Storage.]{
    \begin{minipage}[t]{\linewidth}
    \centering
    \includegraphics[width=\textwidth]{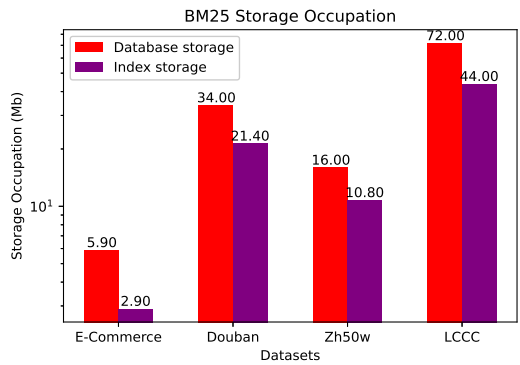}
    \end{minipage}%
  }%

  \subfigure[Dense Storage.]{
    \begin{minipage}[t]{\linewidth}
    \centering
    \includegraphics[width=\textwidth]{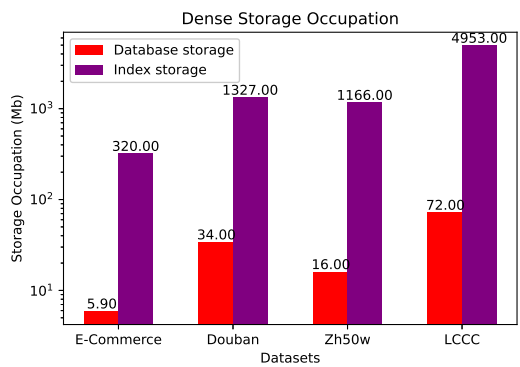}
    \end{minipage}
  }%

  \subfigure[DSHC-128 Storage.]{
    \begin{minipage}[t]{\linewidth}
    \centering
    \includegraphics[width=\textwidth]{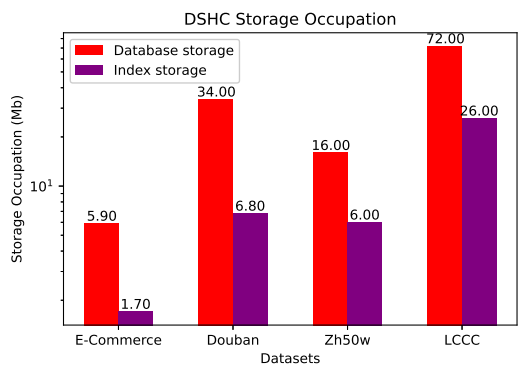}
    \end{minipage}
  }%
  \centering
  \caption{The index and pre-constructed database storage occpuation in four datasets.
  It should be noted that the scale is nonlinear.}
  \label{img:3}
\end{figure}

The stoarge occupation of the pre-constructed database and index are shown in Figure \ref{img:3}.

\end{document}